\pdfoutput=1
\documentclass[runningheads]{article}

\PassOptionsToPackage{numbers, compress}{natbib}
\usepackage[preprint]{neurips_2022}

\usepackage[utf8]{inputenc} % allow utf-8 input
\usepackage[T1]{fontenc}    % use 8-bit T1 fonts
\usepackage{hyperref}       % hyperlinks
\usepackage{url}            % simple URL typesetting
\usepackage{booktabs}       % professional-quality tables
\usepackage{amsfonts}       % blackboard math symbols
\usepackage{nicefrac}       % compact symbols for 1/2, etc.
\usepackage{microtype}      % microtypography
\usepackage{xcolor}         % colors
\usepackage{graphicx}
\usepackage{subcaption}
\usepackage{stackengine}

\newcommand{\vect}[1]{{{\bf #1}}}

\newcommand{\restr}[2]{{% we make the whole thing an ordinary symbol
  \left.\kern-\nulldelimiterspace % automatically resize the bar with \right
  #1 % the function
  \vphantom{\big|} % pretend it's a little taller at normal size
  \right|_{#2} % this is the delimiter
  }}

\title{Long Scale Error Control in Low Light Image and Video Enhancement Using Equivariance}

\author{%
  Sara Aghajanzadeh \\
  Department of Computer Science\\
  University of Illinois Urbana Champaign\\
  \texttt{saraa5@illinois.edu} \\
  \And
  David Forsyth \\
  Department of Computer Science\\
  University of Illinois Urbana Champaign\\
  \texttt{daf@illinois.edu} \\
}

\begin{document}
\maketitle
\begin{abstract}

Image frames obtained in darkness are special. Just multiplying by
a constant doesn't restore the image.   Shot
noise, quantization effects and camera non-linearities
mean that colors and relative light levels are estimated poorly.
Current methods learn a mapping using real dark-bright image
pairs.  These are very hard to capture.  A recent paper has
shown that simulated data pairs produce real improvements
in restoration, likely because huge volumes of simulated data
are easy to obtain.
In this paper, we show that respecting equivariance -- the color
of a restored pixel should be the same, however the image is
cropped -- produces real improvements over the state of the
art for restoration.  We show that a scale selection mechanism
can be used to improve reconstructions.  Finally, we show that
our approach produces improvements on video restoration as
well.  Our methods are evaluated both quantitatively and qualitatively.
\end{abstract}

\section{Introduction}
Image frames obtained in darkness (dark images) are special. Just multiplying by
a constant doesn't restore the image to one taken under a bright light (a bright image).   Shot
noise, quantization effects and camera non-linearities mean that this procedure produces odd
colors and errors in relative brightness.  Current methods multiply the dark image by a gain, then pass
the result through a U-net.  A recent paper~\cite{recent} has shown that simulated data can be used to train this
U-net very effectively, producing significant improvements over methods trained with paired data.
But there are properties of images that the U-net does not capture, and should.

The mapping from a dark image to a bright image has a local character - for a large enough image patch, the bright image
recovered from the patch should be the same as those recovered from the whole image. Moreover, the mapping should be
equivariant under at least affine transformations of the image - for any two images of the same dark scene, the
estimated bright image should be the same in overlapping regions.   Formal equivariance isn't available, because the
boundaries of the image mean that the theory of group actions doesn't apply.  Instead, we use an averaging procedure --
which could be seen as an ensemble estimate -- to impose a weaker, equivariance-like property.  Our approach
obtains a significant improvement in performance over~\cite{recent}.

Combining simulated data and averaging produces results competitive with the state-of-the art on video on the standard
dataset.  Our study of video data exposes an important experimental difficulty with the standard paired test data
set -- all frames are obtained by a camera with the same non-linear camera response.  This means that training on the training set will produce methods
that are notably better on the particular test video, without necessarily being effective on ``in the wild'' dark video
which is likely to have different camera non-linearities.   Our method, if fine-tuned on the training data, is
competitive with state of the art.  If not fine-tuned -- and so capable of dealing with ``in the wild'' video -- it
significantly outperforms natural baselines that are not fine-tuned.

%Show evidence of long scale error, aliasing, and ? with qualitative result that our strategy helps refine.

\section{Background}
\label{sec:related}

A function $\phi: {\bf x} \in X \rightarrow {\bf y} \in Y$ is equivariant under the action of a group $G$ if there
are actions of $G$ on $X$ and $Y$ such that $\phi(g \circ {\bf x})=g \circ \phi({\bf x})$.
An alternative statement of the equivariance property will be convenient.  
Equivariance means that we can choose a convenient coordinate system in which to evaluate $\Phi(f)$ at $\vect{p}$.
We have that, for {\em any} $g \in G$, 
\[
  (g^{-1}\circ \Phi \circ g) (f) (\vect{p})
\]
does not depend on $g$.  In turn, this supplies a formal construction of an equivariant operation $\Psi_{\mbox{eq}}$ out of any operation $\Psi$: we could simply average over $G$, to have
\[
  \Psi_{\mbox{eq}}(f)=\left[\int_{g\in G} (g^{-1} \circ \Psi \circ g)(f) dg\right]/\left[\int_{g \in G} dg\right],
\]
assuming that the integrals can be constructed, etc.  Unfortunately, for most group actions of interest there are very
few equivariant mappings that we can evaluate in practice, so there is no reason to construct the integral. If the
mapping is per pixel -- for example, $\Phi: I(x, y)\rightarrow  I^2(x, y)$ -- it is equivariant, but such mappings are
seldom of interest.  For other mappings, evaluating $\Phi(f)$ at the point $u, v$ requires knowing $f$ in some window
${\cal S}_{u, v}$ that depends on $u, v$ and is larger than a single pixel. Because we know the image only within some
viewport on the image plane, we cannot evaluate the mapping for any $u, v$ such that any part of ${\cal S}_{u, v}$ lies
outside the viewport. Avoiding this problem (for example, by modelling the image as a function on the torus or working
with complete spherical images) leads to a rich theory rooted in harmonic analysis~\cite{HA,GD}.
Padding the image is not a solution, because padding means that the process used to evaluate
$\Phi(f)$ for $u, v$ close to the boundary is different from that for $u, v$ near the center.
Further, the problem can be avoided for some finite group actions~\cite{CohenWelling}, and there is good evidence that
well-known feature representations are approximately equivariant~\cite{LencVedaldi}.

  There is good evidence that imposing equivariance properties improves models. 
Imposing permutation equivariance results in better performing learned set-to-set mappings~\cite{PermInv}.  Functions of
point clouds can be equivariant, and~\cite{EGN} show performance improvements from an E(n)  equivariant construction of
a graph neural network on point clouds.  An E(3) equivariant construction for neural interatomic potentials appears
in~\cite{EPot}.  A general theory for graph neural networks is in~\cite{ET}. Ignoring equivariance considerations in
image-to-image mapping because the theory of group actions doesn't apply is unwise.  For many very interesting
image-to-image mappings, the estimate at a pixel should not depend on where the pixel is in the image.  For 
example, if $\Phi$ maps images to albedos, then the albedo depends on the physical object
being viewed, so that if -- say -- we move the viewport to the left, the albedo should move to the right but not
otherwise change. \cite{DAFRockTPAMI} show that a simpler version of our averaging construction produces significant improvements in albedo estimates.

\section{Methodology}
\label{sec:method}
\subsection{Enhancement model}
We wish to take a dark image and a gain constant, and produce the corresponding bright image.  We achieve this
by multiplying the dark image by the gain constant, then applying a U-net with a novel averaging procedure to the
result.  The final output should be the bright image.  This procedure should work for
{\em any} dark image (``in the wild''), meaning that we cannot expect to know the camera response function (the mapping
from sensor values to pixel values, which is never linear for consumer cameras) for the camera that obtained the image.
The camera response function is important, because known noise models apply to sensor values not pixel values.
An important obstacle is that true paired data -- aligned images of the same scene under different lightings, collected
with different camera response functions -- is extremely hard to collect.

{\bf Simulated training data:} Following~\cite{recent}, we use a straightforward simulation of the imaging pipeline to generate
representative and diverse data.  For the
convenience of the reader, we briefly describe the simulation procedure (~\cite{recent} for more details).   
We take arbitrary normal light images and pass them through a randomly chosen inverse camera response function to linearize. We multiply the linearized image by a small randomly chosen weight to make it dark, then simulate a real
low light image by adding random shot and read noise, then pass it through another randomly chosen camera response function. The procedure yields a
very large simulated paired training dataset. Further, the model input and output are dark and bright images processed
in LAB color space as it is shown to be more effective than RGB space.  

{\bf Model and implementation details:} Following~\cite{recent} we use a U-net~\cite{unet_2015} with
Resnet18~\cite{resnet_2015} backbone generator and a patch discriminator~\cite{im2im_2018}. Similarly, we use four
losses for supervision including L1 loss, L2 loss, and perceptual loss to train the generator for 8 epochs, and then L1
loss and adversarial loss to train the generator and discriminator for additional 8 epochs. 

One of our contributions is applying such model with the effective dataset simulation strategy in video applications
where it is indeed hard to capture aligned dark and bright video frames. We show the performance of the proposed
model~\cite{recent} is not enough on dynamic video data and can be improved on static image data, so we propose
additional techniques to improve the performance.  
\subsection{Equivariance and averaging}

We wish to model an image-to-image mapping that we expect naturally has an equivariance property under
some group $G$ which acts on the image plane (for example, a map from image to albedo, or from dark
image to bright image, should be equivariant under at least rotation, translation and scale).
The workhorse of image-to-image mapping is the U-net~\cite{unet_2015},
an image mapper that is flexible as to the size of the input image.
U-nets are defined on sampled images.  A U-net will not accept an image sampled on a grid with too few
samples, because the subsampling processes in the encoder will produce a data block that is empty.

\begin{figure}
  \centering
  \includegraphics[width=5in]{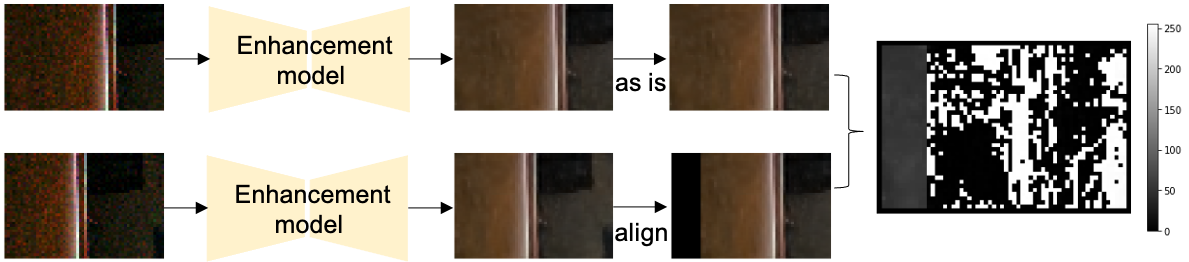}
  \caption{An example where estimations from U-net are different in overlapping region of two crops. The gray scale image shows the absolute difference between image intensities where the overlapping region would theoretically be equal to zero. It is not because the context used to produce the estimates is different from crop to crop.}
  \label{fig:crop-diff}
\end{figure}

Without loss of generality, we choose some $D$ and always apply our U-net to a $D\times D$ grid (a {\em tile}).
We model an image as a function on the unit square $\mathbb{U}=[0, 1]\times [0, 1]$ and a U-net as
an object that will map any image tile, sampled on a $D \times D$ grid, to another function
defined on a $D \times D$ grid.  
Assume we have trained a U-net to implement this mapping in the usual way; write $\Phi_U$ to represent this U-net.
There is no prospect that the U-net will actually be equivariant, because
training procedures do not impose equivariance; the architecture does not guarantee it; and interesting mappings that
are formally equivariant are not available anyhow.    Fig.~\ref{fig:crop-diff} illustrates an example where overlapping
crops given to the UNet model result in different estimations in the overlapping region. 

However, a relaxed version of the procedure to obtain an equivariant mapping from any mapping is extremely interesting.
Write $S$ for the {\em s}ampling operator that maps a function on $\mathbb{U}$
to sampled version of that function on a $D\times D$ and $R$ for a {\em r}econstruction operator that maps a $D\times D$
sampled grid to a continuous function on $\mathbb{U}$.  Write ${\cal R}_{\vect{p}}=
\left\{ g \in G | g^{-1}(\mathbb{U}) \in \mathbb{U} \& g(\vect{p}) \in \mathbb{U}\right\}$ --
for the set of group operations that takes some window $\vect{p} \ni W$ in $\mathbb{U}$ to $\mathbb{U}$.
We consider
\[
  \small
  \Phi_{u, \mbox{eq}}(f)(\vect{p})=\left[\left(\int_{g \in {\cal R}_\vect{p}} w(g) (g^{-1} \circ R \circ \Phi_u \circ
    g)(\restr{f}{g^{-1}(\mathbb{U})})(\vect{p})\right) dg \right]/\left[\int_{g \in {\cal R}_\vect{p}} w(g) dg\right]
\]
Here $w(g)$ is a weighting function; for the moment, assume this is one everywhere.
Notice this does not result in an equivariant mapping because we cannot average over all group operations -- the ones
that lead to windows outside $\mathbb{U}$ are omitted.  Furthermore, this averaging process is not meaningful if
the mapping we are trying to model is not equivariant, because then averaging over $G$ or parts of it is not helpful.
The averaging process has important and interesting properties. The estimate of the mapped value at location $\vect{p}$
is an ensemble estimate obtained by averaging over many different estimators
\[
\Phi_{u, g}(f)(\vect{p})=\left[(g^{-1} \circ R \circ \Phi_u \circ g)(\restr{f}{g^{-1}(\mathbb{U})})\right](\vect{p})
  \]
  (which estimates the value of the mapped $f$ at point $\vect{p}$).  The ensemble estimate
  may have reduced variance. The estimators are different, because the U-net sees a different image window for each $g$
  in the average.  However, training practices mean the estimators should have zero mean (where the random element is
  the choice of window).  

The U-net will be trained with a large number of distinct image crops, and the loss will require that each predicted
value be close to the true value.  Assuming that the training data is extremely large, the U-net will have seen many
distinct windows surrounding a particular pixel, and will be trained to predict the same value for each.  The random
element of the estimate at a particular pixel is the choice of window containing that pixel that is presented to the
U-net.  We can expect that training will result in a U-net that has zero mean error.

Zero mean error at each pixel is not the same as error that has no spatial structure.  We expect that the error at
different locations in the output of the U-net is correlated over some range of scales, because many pairs of output
units have overlapping receptive fields.  This means the error could take the form of a moderately sized,
spatially slow, but structured, error field (Fig.~\ref{fig:crop-diff}).  

An ensemble estimate can control this class of error if we can force down the variance at each location.  This occurs
if the error produced by each of the estimators $\Phi_{u, g}(f)(\vect{p})$ in the average is ``sufficiently
independent'' and if we do not average in estimators with large variance.  As Sec~\ref{sec:ensemble} demonstrates, this can be
achieved in practice.   If we have some reliable method of identifying estimators with large variance, the weighting
function can be used to down weight them.  As Sec.~\ref{sec:ensemble} demonstrates, this can be achieved in practice.

A network should not change prediction if the input image is shifted or scaled.
In other words, an ideal method will report the same estimation for the same location in a scene, however that location is viewed. We know of no crisp theoretical framework to impose this criterion. The theory of group actions does not exactly apply to transformations of the input image such as shifting, cropping, scaling or even rotating because almost all transformations of this form involve information being gained or lost at the boundary of the image~\cite{DAFRockTPAMI}. 

%Augmentation and averaging result in a property analogous to invariance, though a precise definition remains obscure.

\begin{figure*}[!ht]
	\centering
	\begin{subfigure}{0.24\textwidth}
		\includegraphics[width=\textwidth]{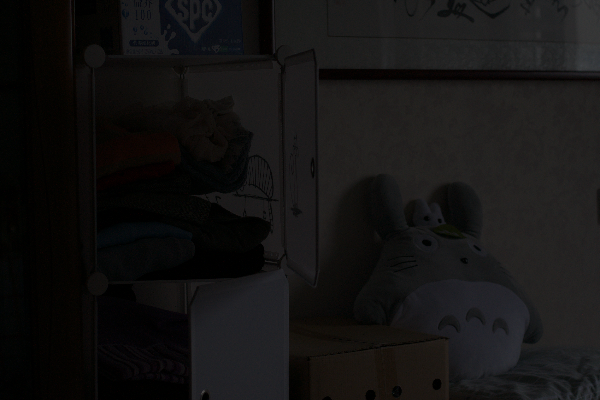}
		\caption{Input}
	\end{subfigure}
	\begin{subfigure}{0.24\textwidth}
		\includegraphics[width=\textwidth]{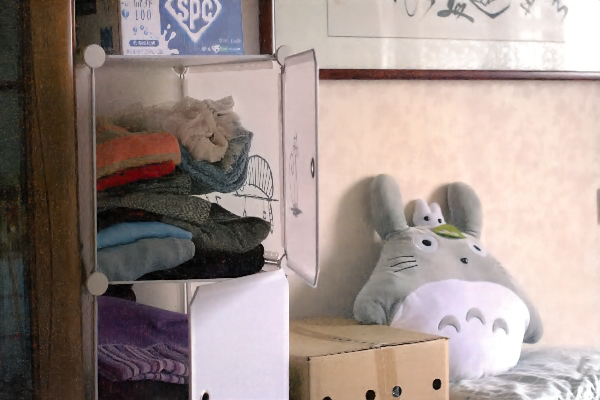}
		\caption{Long scale restoration}
	\end{subfigure}
	\begin{subfigure}{0.24\textwidth}
		\includegraphics[width=\textwidth]{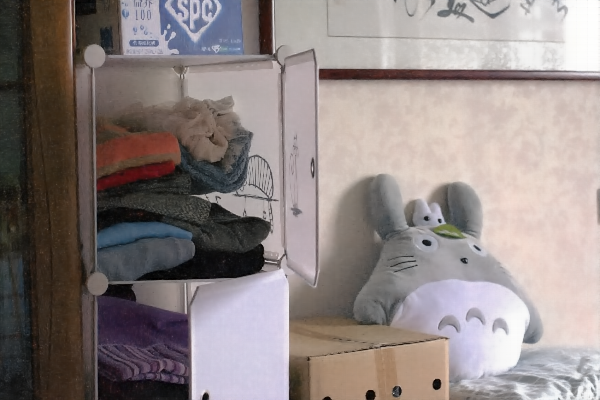}
		\caption{Short scale restoration}
	\end{subfigure}
	\begin{subfigure}{0.24\textwidth}
		\includegraphics[width=\textwidth]{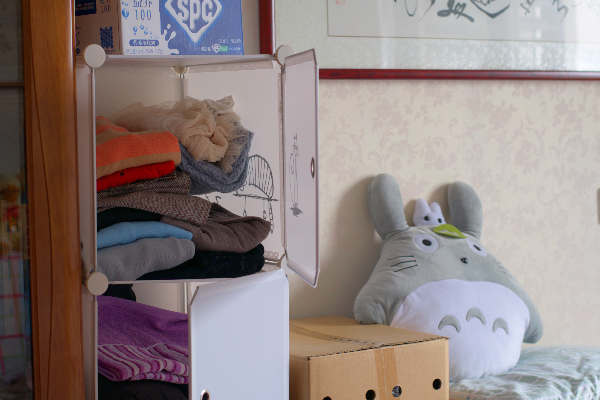}
		\caption{Ground truth}
	\end{subfigure}
	\begin{subfigure}{0.24\textwidth}
		\includegraphics[width=\textwidth]{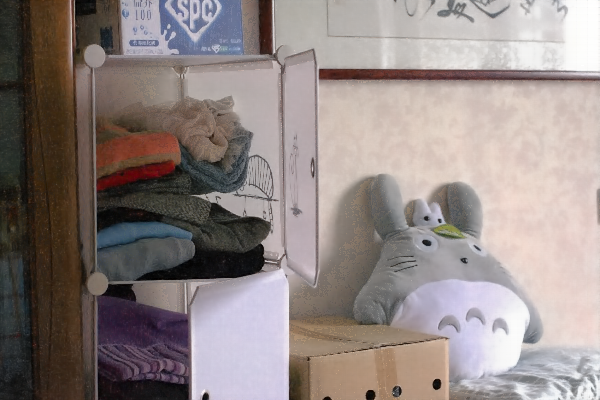}
		\caption{Oracle Image}
	\end{subfigure}
	\begin{subfigure}{0.24\textwidth}
		\includegraphics[width=\textwidth]{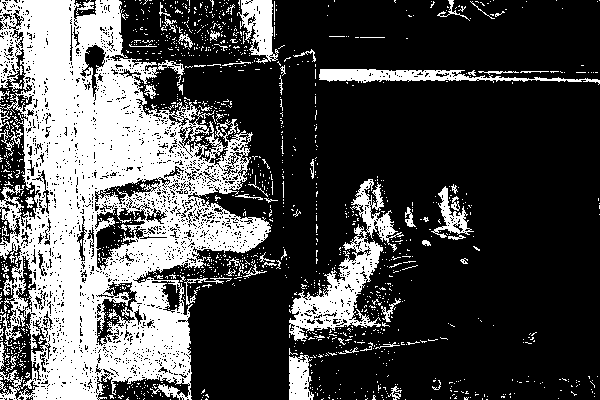}
		\caption{Oracle mask}
	\end{subfigure}
	\begin{subfigure}{0.24\textwidth}
		\includegraphics[width=\textwidth]{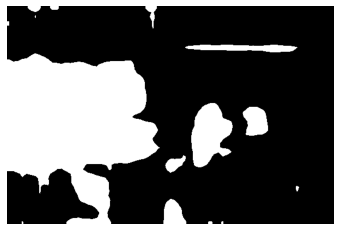}
		\caption{Scale prediction}
	\end{subfigure}
	\begin{subfigure}{0.24\textwidth}
		\includegraphics[width=\textwidth]{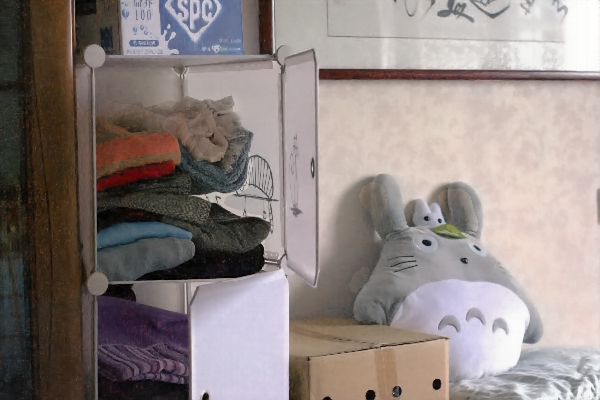}
		\caption{Ensemble estimate}
	\end{subfigure}
	\caption{A sample image from LOL~\cite{LOL_2018} test-set\label{fig:lol22-method}}
\end{figure*}

\textbf{Weighting the windows.}
Choosing $w(g)$ correctly is important.
The average must be sampled, and preliminary experiments yield that the more tiles overlap a pixel, the better the
final estimation is.  However, there are diminishing returns and the inference time grows with the number of tiles.  We
cover the image with a grid of randomly offset tiles that overlap the next tile by 80\%.  Furthermore, the size of the
tile used at a particular location matters. If the tile is too small a fraction of the image, the prediction is poor,
likely because for particularly dark image patches, local context is important.  Similarly, if the tile is too large a
fraction of the image, the prediction is poor, likely because for brighter patches, texture effects mean that using too
much context creates errors. Fig.~\ref{fig:lol22-method} illustrates an example where the first
row (a)-(d) show the dark image input, model estimate at long scale and short scale, and the bright ground truth image
respectively.

%We show how to learn automatic scale selection for our ensemble estimate. Along with the averaging procedure, the scale selection network can estimate which scale to be used where in the image in order to produce the best possible output. This scale selection model can learn from data.

% \textbf{equivariance time vs. no equivariance time}.
%Though we find our averaging procedure to impose equivariance helps control long scale error as well.

\subsection{Ensemble estimate}
\label{sec:ensemble}
\textbf{Scale selection.}
% bunny/cabinet figure shows oracle label map to be used for supervising the scale selection network. 
We find that in some places, a very short scale estimate is the way to go. In other places, it must be the long estimate. In other words, the ideal is to use large tiles in dark patches and small tiles in bright patches. This can be achieved by manipulating
$w(g)$.
We introduce the scale selection model which is an off-the-shelf semantic segmentation DeepLabv3~\cite{deeplabv3} model for the task of scale classification using transfer learning. 
%Segmentation is considered a dense classification task as every pixel is classified into a predefined class.  
The masks are class indexed masks, an example shown in Fig.~\ref{fig:lol22-method} (f).
There are two classes, class 0 indicating short scale and class 1 indicating long scale. 

%So if a particular pixel is class 0, it comes from the estimate at long scale (whole image fed to the model without tiling strategy) and if it is class 1, it comes from the estimate at short scale (small tiles fed to the model to reconstruct the output with tiling strategy). 

Although the scale selection model and the enhancement model get the same input image, they are not trained end-to-end because it is hard to tell the network to have a big receptive field in some regions but small one in other regions. 
%We use the well known categorical cross entropy loss function where each pixel can belong to only one class. (or mse loss when binary) / how many epochs?
To supervise the scale selection model, we produce the train-set and test-set (sample shown in Fig.~\ref{fig:lol22-method} (e)). To create the label masks (sample shown in Fig.~\ref{fig:lol22-method} (f)), we take the ground truth image and choose the pixel estimate that's closest to the ground truth at each location.
This way, we can guide the scale selection model to learn where in the image should be short scale estimate and where should be long scale estimate. The reconstructions we get and the scores we achieve are an upper bound to what our scale selection model would actually achieve, so we call it the oracle. The oracle scores tell us how much we benefit from scale selection mechanism.

\begin{figure}
  \centering
  \includegraphics[width=5.5in]{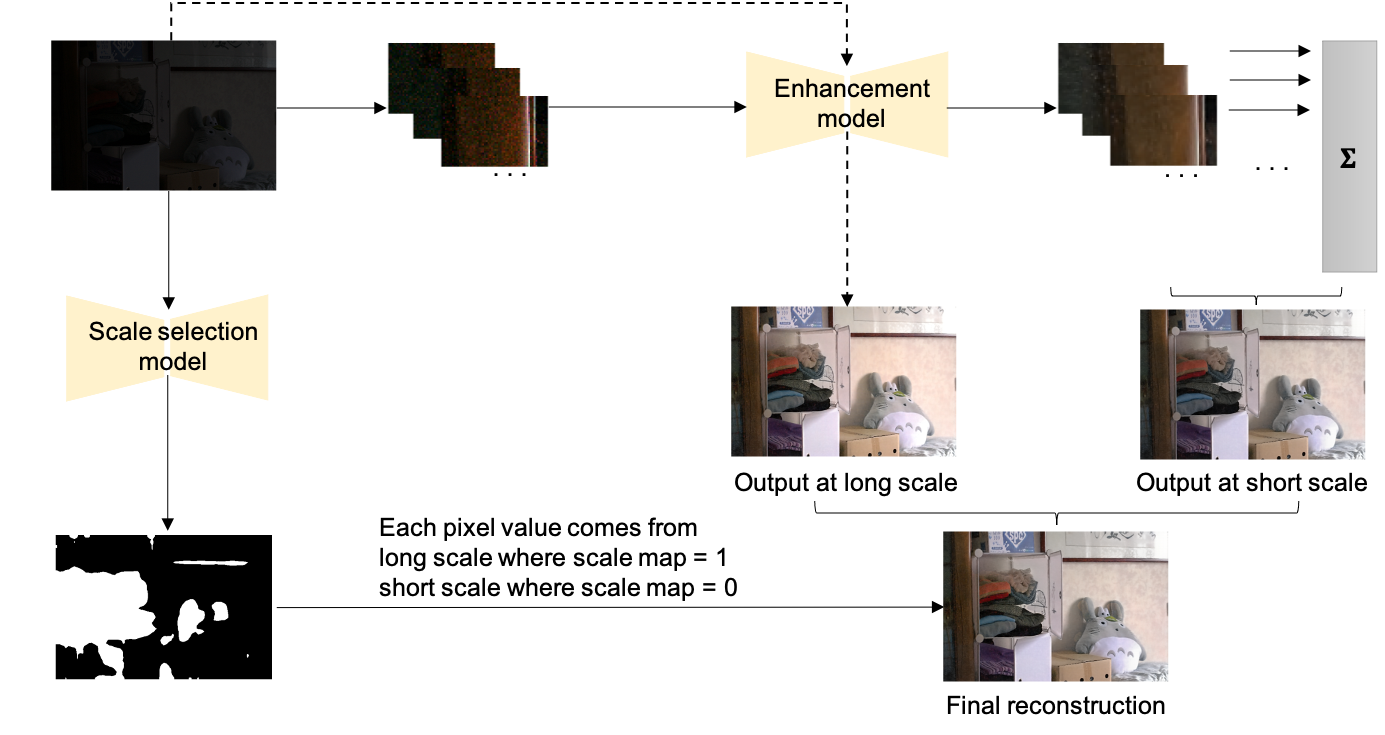}
  \caption{Proposed methodology at inference: our enhancement model receives a dark input image and predicts a bright output image at long scale. It also receives the input image crops and reconstructs the predicted output image at short scale by the averaging procedure introduced in this paper. Moreover, the scale selection model receives the same input image as our enhancement model and predicts the scale map. To contruct the final output image, the scale map is used to guide the pixel value selection for the ensemble estimate.}
  \label{fig:method}
\end{figure}

We now bring the enhancement model and automatic scale selection together to get our ensemble estimate. Fig.~\ref{fig:method} shows the overall pipeline.
We take the enhancement model to estimate at short and long scales. We have a trained scale selection model that accepts the same input image as the enhancement model, and gives us the scale map prediction (sample shown in Fig.~\ref{fig:lol22-method} (g)) that we use to estimate the pixel values.

%There is some natural scale for the process. Intuitively, averaging won't help if the tile size is too small as the variance is very high. The plot shows that the best smallest tile size we can use is at 10\% coverage of the whole image. Thus, if an image is 512x341, the tile size we could use is 51x34. Smaller than that and the error starts increasing, forcing the PSNR to go down. 

%Another experiment we did shows that the second best scale is fairly close to the first in terms of metric score, suggesting that mild inefficiency if any is not the worst thing in the world. Further, we can smooth multiple scales to have even better estimates as well as smooth qual res. 

%For SDSD, our final numbers are from 3 classes, short, long, or average of both. because of bright delicate structure on dark region; high spatial frequency regions. 

%If good, proceed with video data: talk about SDSD; why not 3DUNet? etc. 

%Can we show temporal consistency for video NOW?

\section{Experimental results}
\label{sec:eval}

\subsection{Static images}

Quantitatively comparison with the state-of-the-art in low light image enhancement strongly supports the
use of our averaged estimate.  As Tab.~\ref{tab:mit-lol} shows, using our average estimate results in an approximate 1 PSNR improvement
over past methods.

\textbf{Datasets.} We use the two paired standard datasets used by~\cite{recent}, LOL~\cite{LOL_2018} test-set and MIT-Adobe-FiveK~\cite{fivek}
test-set. Please note our method is not trained or fine-tuned on these two datasets, but uses only simulated data.
The entire LOL dataset contains 500 static color images at 400x600 resolutions. The test-set contains only 15
images. The MIT-Adobe-FiveK dataset contains 5000 images exported in PNG format. The test-set contains only 500 images.   

\textbf{Quantitative results.}
We use the two widely used reference-based image quality metrics, peak-signal-noise-ratio (PSNR) and structural-similarity-index-measure (SSIM). We compare the proposed method with representative baselines. 
%We rely on benchmarking results provided in the recent paper~\cite{recent} and use them for our comparisons. 
Our ensemble estimate - averaging procedure and automatic scale selection - shows significant improvement on restoring static images.
Tab.~\ref{tab:mit-lol} shows that our proposed method outperforms state of the art methods by a great margin on LOL test-set and  MIT test-set in terms of PSNR. MBLLEN~\cite{MBLLEN_2018} achieves the best SSIM score, and DSLR~\cite{DSLR_2021} achieves the second best SSIM score. Please note that it is trained on the MIT train-set but ours is not because we show the generalization capability of our model.

\begin{figure}
\stackunder[1pt]{\includegraphics[width=0.24\textwidth]{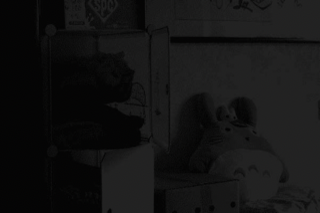}}{Input}
\stackunder[1pt]{\includegraphics[width=0.24\textwidth]{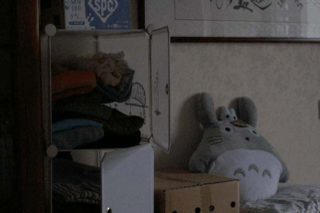}}{RRDNet~\cite{RRDNet_2020}}
\stackunder[1pt]{\includegraphics[width=0.24\textwidth]{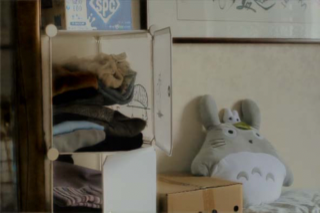}}{DRBN~\cite{DRBN_2020}}
\stackunder[1pt]{\includegraphics[width=0.24\textwidth]{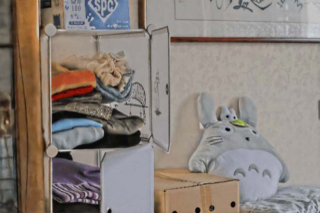}}{KinD++~\cite{KinD++_2021}}
\stackunder[1pt]{\includegraphics[width=0.24\textwidth]{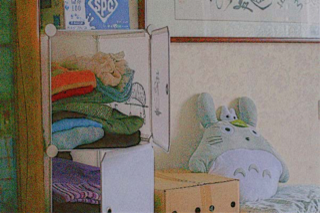}}{RetinexNet~\cite{Retinex-Net_2018}}
\stackunder[1pt]{\includegraphics[width=0.24\textwidth]{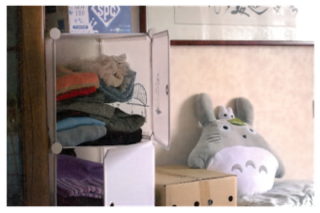}}{Recent work~\cite{recent}}
\stackunder[1pt]{\includegraphics[width=0.24\textwidth]{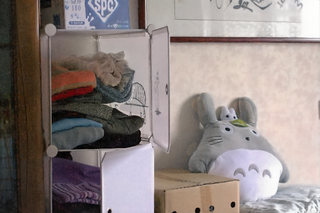}}{Ours}
\stackunder[1pt]{\includegraphics[width=0.24\textwidth]{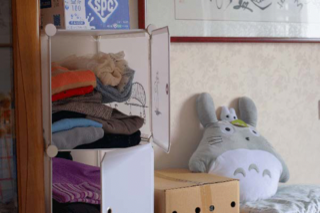}}{Ground truth}
\stackunder[1pt]{\includegraphics[width=0.24\textwidth]{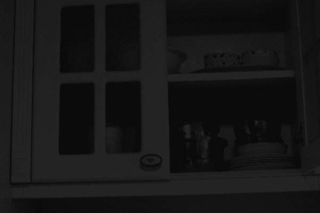}}{Input}
\stackunder[1pt]{\includegraphics[width=0.24\textwidth]{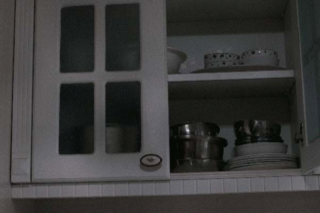}}{LightenNet~\cite{LightenNet_2018}}
\stackunder[1pt]{\includegraphics[width=0.24\textwidth]{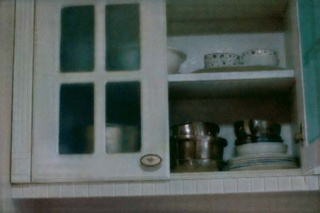}}{DSLR~\cite{DSLR_2021}}
\stackunder[1pt]{\includegraphics[width=0.24\textwidth]{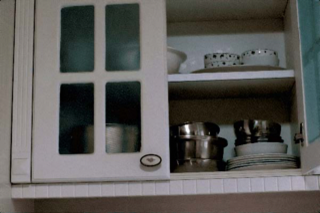}}{MBLLEN~\cite{MBLLEN_2018}}
\stackunder[1pt]{\includegraphics[width=0.24\textwidth]{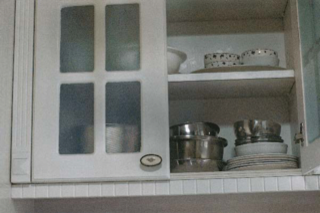}}{EnlightenGAN~\cite{EnlightenGAN_2021}}
\stackunder[1pt]{\includegraphics[width=0.24\textwidth]{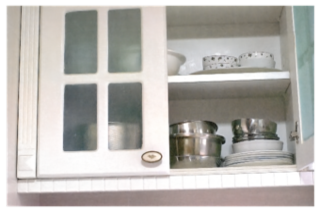}}{Recent work~\cite{recent}}
\stackunder[1pt]{\includegraphics[width=0.24\textwidth]{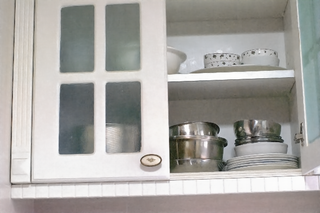}}{Ours}
\stackunder[1pt]{\includegraphics[width=0.24\textwidth]{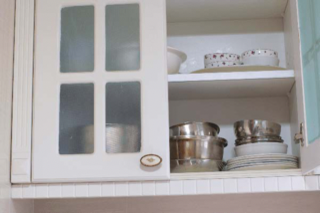}}{Ground truth}
\stackunder[1pt]{\includegraphics[width=0.24\textwidth]{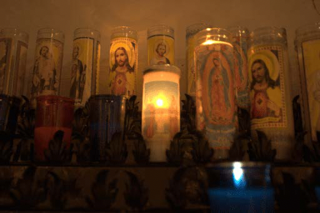}}{Input}
\stackunder[1pt]{\includegraphics[width=0.24\textwidth]{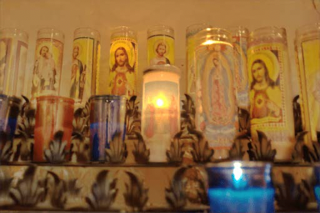}}{KinD~\cite{KinD_2019}}
\stackunder[1pt]{\includegraphics[width=0.24\textwidth]{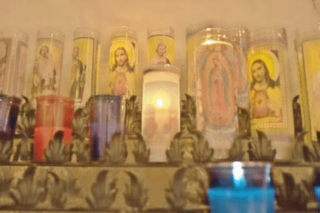}}{LLNet~\cite{LLNet_2017}}
\stackunder[1pt]{\includegraphics[width=0.24\textwidth]{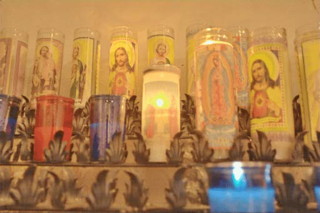}}{TBEFN~\cite{TBEFN_2021}}
\stackunder[1pt]{\includegraphics[width=0.24\textwidth]{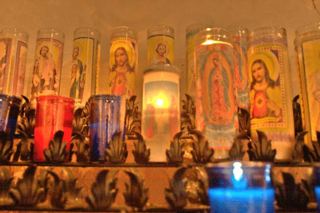}}{ExCNet~\cite{ExCNet_2019}}
\stackunder[1pt]{\includegraphics[width=0.24\textwidth]{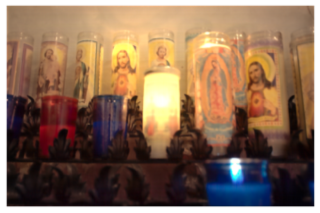}}{Recent work~\cite{recent}}
\stackunder[1pt]{\includegraphics[width=0.24\textwidth]{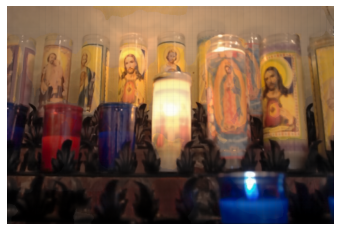}}{Ours}
\stackunder[1pt]{\includegraphics[width=0.24\textwidth]{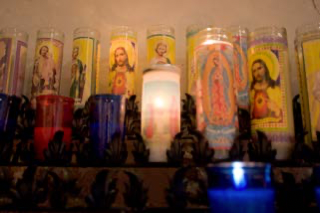}}{Ground truth}
\stackunder[1pt]{\includegraphics[width=0.24\textwidth]{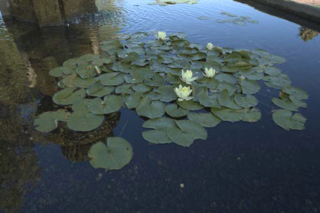}}{Input}
\stackunder[1pt]{\includegraphics[width=0.24\textwidth]{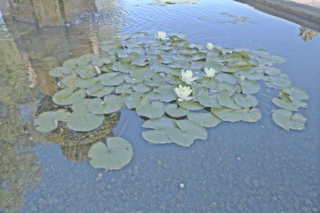}}{ZeroDCE~\cite{Zero-DCE_2019}}
\stackunder[1pt]{\includegraphics[width=0.24\textwidth]{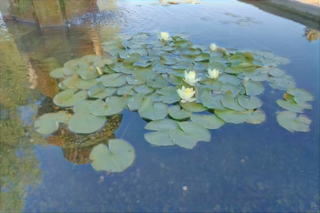}}{DRBN~\cite{DRBN_2020}}
\stackunder[1pt]{\includegraphics[width=0.24\textwidth]{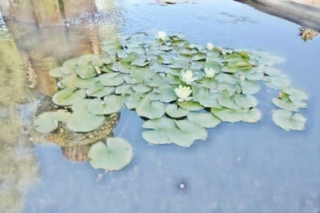}}{KinD++~\cite{KinD_2019}}
\stackunder[1pt]{\includegraphics[width=0.24\textwidth]{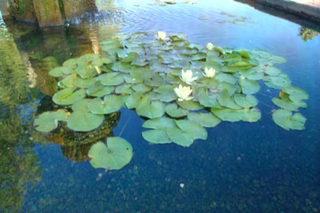}}{DSLR~\cite{DSLR_2021}}
\stackunder[1pt]{\includegraphics[width=0.24\textwidth]{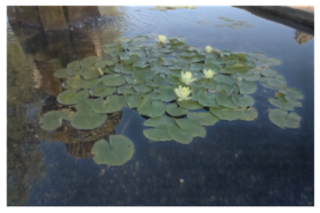}}{Recent work~\cite{recent}}
\stackunder[1pt]{\includegraphics[width=0.24\textwidth]{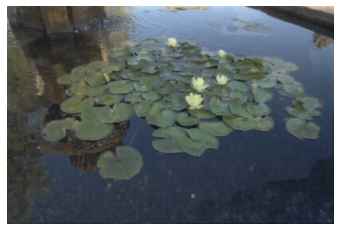}}{Ours}
\stackunder[1pt]{\includegraphics[width=0.24\textwidth]{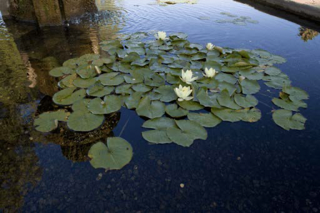}}{Ground truth}
\caption{Results on images from LOL (top two) and MIT-Adobe-FiveK (bottom two)}
\label{fig:qual-static}
\end{figure}

\textbf{Qualitative results.}
 Fig.~\ref{fig:qual-static} shows some qualitative results on the LOL and MIT-Adobe-FiveK test-sets. As shown, our results are fairly close to the ground truth image. For the first example, the overall brightness is similar, but the color of the wood shown in the left is a bit darker in ours versus the ground truth. None of the methods correctly recover the color of the violet piece of clothing in the left. For the second example, other baselines do not improve the brightness of the low light input image very well whereas our predicted output is similar to the ground truth image. 
 For the third example, other baselines improve brightness of the low light input image but they produce either overexposed or underexposed results whereas our predicted output closely matches the ground truth image. Similarly for the last example, our method produces the closest bright image to the ground truth image.

\begin{table}
  \caption{Quantitative comparisons in terms of reference-based metrics (PSNR (in dB), SSIM) among our method and state-of-the-art baselines on static images. We did not use these datasets at train time. There is a check mark indicating whether other methods use them at train time.}
  \label{tab:mit-lol}
  \centering
  \resizebox{\textwidth}{!}{\begin{tabular}{lcllcll}
  %\begin{tabular}{lllllll}
    \toprule
    %\multicolumn{3}{c}{MIT Dataset}{3}{c}{lol Dataset}                \\
    \multicolumn{1}{c}{} &
    \multicolumn{3}{c}{MIT Dataset} &
    \multicolumn{3}{c}{LOL Dataset} \\
    \cmidrule(r){1-7}
    Methods     & trained on MIT? & PSNR     & SSIM  & trained on LOL? & PSNR & SSIM\\
    \midrule
    LLNet~\cite{LLNet_2017} &  & 12.177  & 0.645     & \checkmark & \textcolor{green}{17.959} & 0.713 \\
    LightenNet~\cite{LightenNet_2018} &  & 13.579 & 0.744 & & 10.301 & 0.402 \\
    Retinex-Net~\cite{Retinex-Net_2018} &  & 12.310 & 0.671 & \checkmark & 16.774 & 0.462 \\
    MBLLEN~\cite{MBLLEN_2018} &  & \textcolor{green}{19.781} & \textbf{0.825} & & 17.902 & 0.715 \\
    KinD~\cite{KinD_2019} &  & 14.535 & 0.741 & \checkmark & 17.648 & 0.779 \\
    KinD++~\cite{KinD++_2021} &  & 9.732 & 0.568 & \checkmark & 17.752 & 0.760 \\
    TBEFN~\cite{TBEFN_2021} &  & 12.769 & 0.704 & \checkmark & 17.351 & \textcolor{green}{0.786} \\
    DSLR~\cite{DSLR_2021} & \checkmark & 16.632 & \textcolor{blue}{0.782} & & 15.050 & 0.597 \\
    EnlightenGAN~\cite{EnlightenGAN_2021} &  & 13.260 & 0.745 & & 17.483 & 0.677 \\
    DRBN~\cite{DRBN_2020} &  & 13.355 & 0.378 & \checkmark & 15.125 & 0.472 \\
    ExCNet~\cite{ExCNet_2019} &  & 13.978 & 0.710 & & 15.783 & 0.515 \\
    Zero-DCE~\cite{Zero-DCE_2019} &  & 13.199 & 0.709 & & 14.861 & 0.589 \\
    RRDNet~\cite{RRDNet_2020} &  & 10.135 & 0.620 & & 11.392 & 0.468 \\
    Recent work~\cite{recent} &  & \textcolor{blue}{19.814} & \textcolor{green}{0.754} & & \textcolor{blue}{20.434} & \textcolor{blue}{0.808} \\ %ours woeq
    \bottomrule
    %Ours w equivariance (fixed short scales) & & 20.447 & &  & 22.973 & 0.820 \\
    Our ensemble estimate (expectation of short and long scales) &  & 20.400 & 0.746 &  & 22.042 & 0.820\\
    Our ensemble estimate (short or long scales) &  & \textbf{21.031} & 0.734 & & \textbf{23.114} & \textbf{0.820}\\
    \textit{Oracle} ensemble estimates &  & 21.532 & 0.762 &  & 23.832 & 0.822\\
    \bottomrule
  %\end{tabular}
  \end{tabular}}
\end{table}

\subsection{Dynamic video frames}
We compare quantitatively and qualitatively with recent works in low light image and video enhancement. Following another recent work~\cite{sdsd}, we adopt the paired SDSD~\cite{sdsd} video dataset. The entire dataset contains 80 outdoor scenes and 70 indoors scenes of which random 13 outdoor scenes and 12 indoor scenes are used for test-set. Following the evaluation protocol~\cite{sdsd}, we use the first 30 frames in each video sequence for the test-set. As such, there are a total of 750 frames in the test-set. 

\textbf{Quantitative results.}
We rely on the benchmarking results presented in the paper~\cite{sdsd} and follow their procedure to compare our method
with the representative baselines in terms of PSNR and SSIM, shown in Tab.~\ref{tab:sdsd}. 
%Please note the other baselines are finetuned on this dataset. Our method does not perform strongly when it is not finetuned on the dataset. In the case of static images, we show that our method generalizes well on unseen datasets because it has seen diverse and representative training data. However, in case of this dynamic video dataset, it has not.
Our scale selected averaging produces significant improvements and achieves the highest SSIM score, but slightly lower
than SMID~\cite{smid} and SDSD~\cite{sdsd} in terms of PSNR.  Tab.~\ref{tab:sdsd} shows standard evaluation protocols
where methods work best on the test split of the dataset they used for training.  

\textbf{Experimental finding.}
\label{exp-finding}
SDSD results must be interpreted with extreme caution.  We find that obtaining a
strong result on SDSD requires our method be fine-tuned on that dataset's training data.  Furthermore,
all static image methods fine-tuned on SDSD display significantly improved PSNR on SDSD (compare 12-19 in
Tab.~\ref{tab:mit-lol} with 20-24 in Tab.~\ref{tab:sdsd}), though one would expect that motion blur means that
training on static images leads to weak performance on video.  This oddity is the result of an important experimental
difficulty with SDSD dataset. All frames, bright and dark, test and train, are obtained by a camera with the same non-linear camera response, because
collecting paired dark-bright data is extremely hard.  This means that methods that do not know this camera response,
either because they are not fine-tuned or because they are agnostic to camera response function, are at a significant
disadvantage on the test-set {\em which is collected with only one camera response function}.
In turn, methods that perform well on SDSD may not necessarily be effective on ``in the wild'' dark video which is
likely to have different camera non-linearities.  Tab.~\ref{tab:sdsd-nofinetune} confirms this effect.  We evaluate a strong baseline
(MBLLEN~\cite{MBLLEN_2018}) on SDSD without fine-tuning; its performance is consistent with performance on static
images.  Similarly, the method of~\cite{recent} without fine-tuning is somewhat better than MBLLEN, as one would expect.
Finally, our averaging method significantly outperforms both.

\begin{figure*}[!ht]
	\centering
	\begin{subfigure}{0.24\textwidth}
		\includegraphics[width=\textwidth]{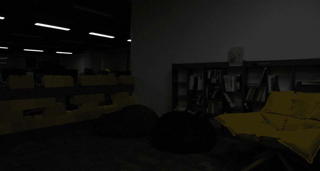}
		\caption{Input}
	\end{subfigure}
	\begin{subfigure}{0.24\textwidth}
		\includegraphics[width=\textwidth]{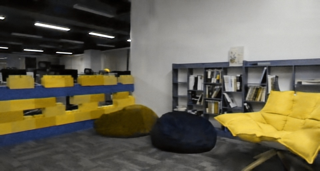}
		\caption{DeepLPF}
	\end{subfigure}
	\begin{subfigure}{0.24\textwidth}
		\includegraphics[width=\textwidth]{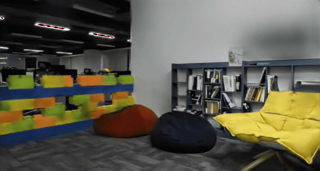}
		\caption{DRBN}
	\end{subfigure}
	\begin{subfigure}{0.24\textwidth}
		\includegraphics[width=\textwidth]{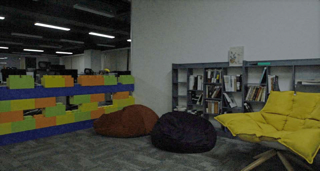}
		\caption{ZeroDCE}
	\end{subfigure}
	\begin{subfigure}{0.24\textwidth}
		\includegraphics[width=\textwidth]{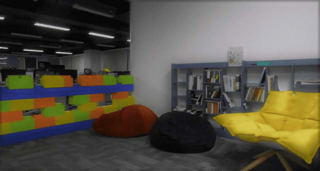}
		\caption{SMOID}
	\end{subfigure}
	\begin{subfigure}{0.24\textwidth}
		\includegraphics[width=\textwidth]{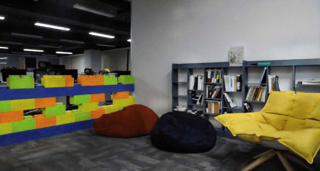}
		\caption{SDSD}
	\end{subfigure}
	\begin{subfigure}{0.24\textwidth}
		\includegraphics[width=\textwidth]{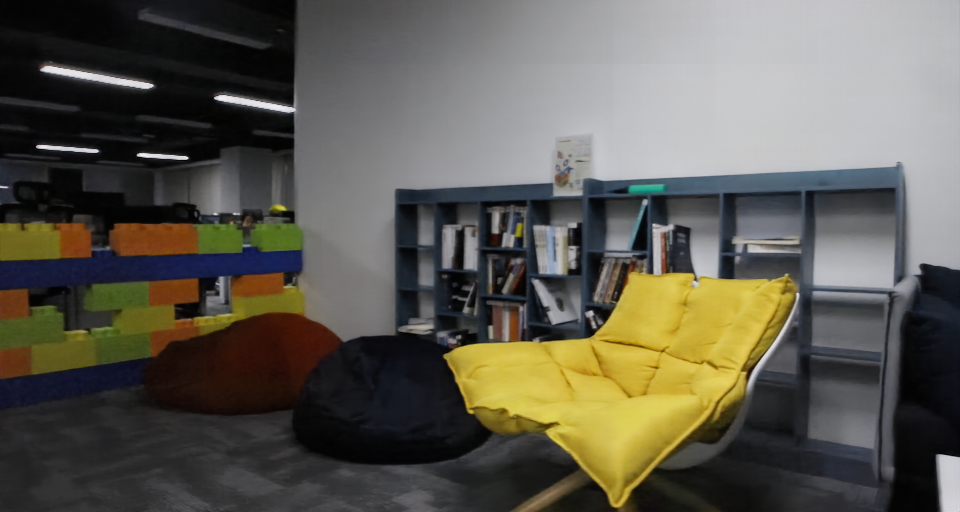}
		\caption{Ours}
	\end{subfigure}
	\begin{subfigure}{0.24\textwidth}
		\includegraphics[width=\textwidth]{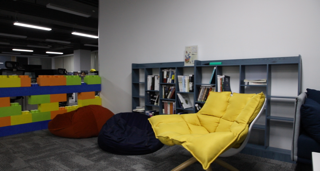}
		\caption{Ground truth}
	\end{subfigure}
	\caption{Qualitative comparison on a sample video frame from SDSD test-set \label{fig:qual-dynamic}}
\end{figure*}

\textbf{Qualitative results.}
 Fig.~\ref{fig:qual-dynamic} shows our qualitative evaluation on a test frame in SDSD tset-set. As shown, our result is fairly close to the ground truth image. For example, our method can retrieve the correct color and brightness on the colorful logos on the left side of the image. Meanwhile, DeepLPF has color deviations, DRBN has shading artifacts, ZeroDCE still recovers a dark output, SMOID result is blurry, and SDSD is somewhat overexposed.

\begin{table}
  \caption{Quantitative comparisons in terms of reference-based metrics (PSNR (in dB), SSIM) among our method and
    state-of-the-art baselines on dynamic video frames. Please note all methods are fine-tuned and refer to
    Sec.~\ref{exp-finding} experimental finding discussion regarding SDSD dataset.}
  \label{tab:sdsd}
  \centering
  %\resizebox{\textwidth}{!}{\begin{tabular}{lrr}
  \begin{tabular}{lrr}
    \toprule
    \multicolumn{3}{c}{SDSD Dataset}                   \\
    \cmidrule(r){1-3}
    Methods     & PSNR     & SSIM  \\
    \midrule
    DeepUPE~\cite{DeepUPE_2019} & 21.82  & 0.68     \\
    ZeroDCE~\cite{Zero-DCE_2019} & 20.06 & 0.61 \\
    DeepLPF~\cite{DeepLPF_2020} & 22.48 & 0.66 \\
    DRBN~\cite{DRBN_2020} & 22.31 & 0.65 \\
    MBLLEN~\cite{MBLLEN_2018} & 21.79 & 0.65 \\
    SMID~\cite{smid} & \textcolor{blue}{24.09} & \textcolor{green}{0.69} \\
    SMOID~\cite{smoid} & 23.45 & 0.69 \\
    SDSD~\cite{sdsd} & \textbf{24.92} & \textcolor{blue}{0.73} \\
    %Ours w/o equivariance (2DUNet) & 18.92 & 0.66 \\
    Recent work~\cite{recent} & 23.16 & 0.79 \\
    \bottomrule
    %Ours w/o equivariance (3DUNet) & 18.47 & 0.68 \\
    %Ours w equivariance (fixed short scales) & 20.10 & 0.69 \\
    %Ours w scale select (short, long, all) & 20.10 & 0.68 \\
    %Ours w scale select (oracle) & 22.45 & 0.76 \\
    %\textbf{After fine-tune weq} & \textbf{20.06} & \textbf{0.66} \\
    %\textbf{After fine-tune woeq} & \textbf{19.91} & \textbf{0.66} \\
    %\textbf{After fine-tune with their pairs-woeq} & \textbf{22.97} & \textbf{0.79} \\
    %\textbf{After fine-tune with their pairs-weq} & \textbf{22.60} & \textbf{0.77} \\
    %below: After additional epochs woeq
    %Our fine-tuned model wo equivariance & 23.16 & 0.79 \\
    Our ensemble estimate (expectation of short and long scales) & 23.38 & 0.80 \\
    Our ensemble estimate (short or long scales) & \textcolor{green}{23.60} & \textbf{0.80} \\
    \textit{Oracle} ensemble estimate & 25.06 & 0.84 \\
    \bottomrule
  \end{tabular}
  %\end{tabular}}
\end{table}

\begin{table}
  \caption{Quantitative comparisons among our method and state-of-the-art baselines \textit{not fine-tuned} on SDSD dataset and refer to Sec.~\ref{exp-finding} experimental finding discussion regarding SDSD dataset.}
  \label{tab:sdsd-nofinetune}
  \centering
  \resizebox{\textwidth}{!}{\begin{tabular}{lrrrr}
    %\begin{tabular}{lrrrr}
    \toprule
    \multicolumn{1}{c}{} &
    \multicolumn{2}{c}{Fine-tuned} &
    \multicolumn{2}{c}{Not fine-tuned} \\
    \cmidrule(r){1-5}
    Methods & PSNR & SSIM  & PSNR & SSIM\\
    \midrule
    MBLLEN~\cite{MBLLEN_2018} & 21.79 & 0.65 & 16.67  & 0.61     \\
    Recent work~\cite{recent} & 23.16 & 0.79 & 18.92 & 0.66 \\
    %^Ours w/o equivariance (2DUNet)
    \bottomrule
    %Ours w/o equivariance (3DUNet) & 18.47 & 0.68 \\
    Our ensemble estimate (short or long scales) & \textbf{23.60} & \textbf{0.80} & \textbf{21.57} & 0.71\\
    Our ensemble estimate (expectation of short and long scales) & 23.38 & 0.80 & 21.21 & \textbf{0.72} \\
    % Ours will be model04 (same model as for mit,lol) and scale_net on sdsd train because that's ok for scales. equivariance hard , equivariance soft/expectation
    \bottomrule
  %\end{tabular}
  \end{tabular}}
\end{table}

\begin{figure}
  \centering
  	\begin{subfigure}{0.24\textwidth}
		\includegraphics[width=\textwidth]{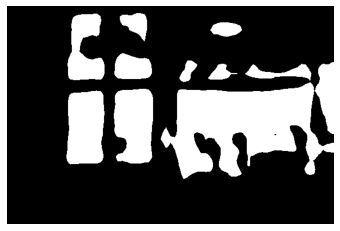}
		\caption{Scale prediction}
	\end{subfigure}
	\begin{subfigure}{0.24\textwidth}
		\includegraphics[width=\textwidth]{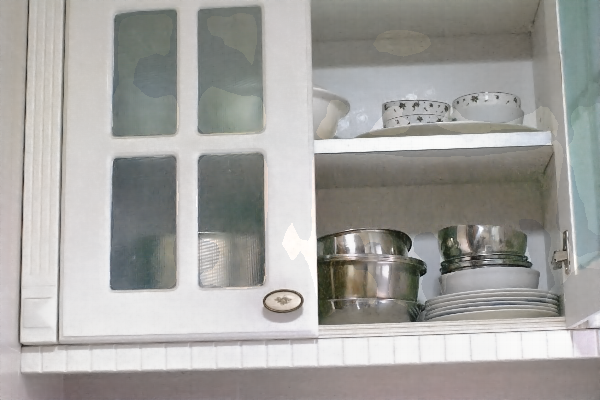}
		\caption{ensemble estimate}
	\end{subfigure}
	\begin{subfigure}{0.24\textwidth}
		\includegraphics[width=\textwidth]{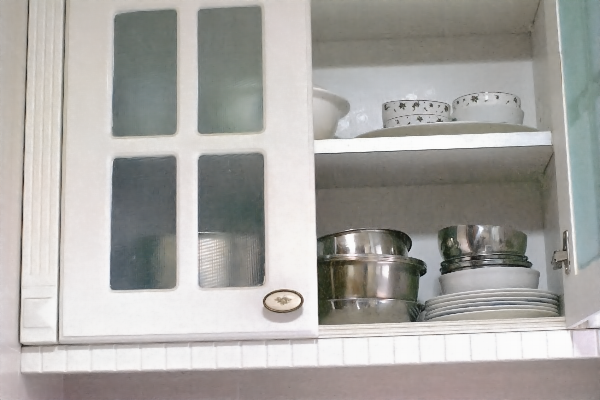}
		\caption{ensemble estimate}
	\end{subfigure}
	\begin{subfigure}{0.24\textwidth}
		\includegraphics[width=\textwidth]{neurips_2022/figs/qual-static/GT3.png}
		\caption{Ground truth}
	\end{subfigure}
  \caption{The scale selection map and final reconstructions of a sample image from LOL~\cite{LOL_2018} test-set. The proposed scale selection model predicts the map shown in (a). To construct the ensemble estimates (b) or (c), two approaches may be taken. (1) Use a hard probability assignment: if model confidently classifies short scale for a particular pixel, then directly assign the pixel value from the short scale estimate; otherwise, from the long scale estimate. Sometimes, this may result in small artifacts. (2) Thus, one may use a soft assignment and compute the expectation of posterior probabilities of short and long scale estimates, and assign the resulting pixel values to construct the ensemble estimate (c). This leads to a better qualitative result without any artifacts.}
  \label{fig:limit}
\end{figure}

 \textbf{Limitation.} The enhancement model estimates at long scale and short scale. Given the scale predictions, one could reconstruct the final image by hard assignment where a short scale estimate is used where scale map indicates class 0 and a long scale estimate is used where scale map indicates class 1. However, this may result in a blocky output image which is not qualitatively pleasing. Consequently, one could produce a visually pleasing output image with a soft assignment or expectation of posterior probabilities predicted by the scale selection model. Although this approach provides a smooth output image, it may result in a slight decrease in PSNR when compared against the ground truth image. An example is shown in Fig.~\ref{fig:limit} where Fig.~\ref{fig:limit}(b) shows a blocky output image with PSNR of $26.713$ and SSIM of $0.889$. Fig.~\ref{fig:limit}(c) shows a smooth output image with PSNR of $24.851$ and SSIM of $0.901$.

\section{Conclusion}
\label{sec:conclusion}
In this paper, we study the problem of brightening dark static images or dynamic video frames found in the wild. We propose a scale selection mechanism to impose equivariance property and averaging technique which could be seen as an ensemble estimate to control long scale errors in image reconstruction.
Our proposed method achieves best in all metrics on a number of standard static image datasets, and very strong on a standard dynamic video dataset. Qualitative comparisons suggest strong improvements over state of the art methods.  

\bibliographystyle{IEEEtran}
\bibliography{main-daf-1.bib}

% Generated by IEEEtran.bst, version: 1.14 (2015/08/26)
\begin{thebibliography}{10}
\providecommand{\url}[1]{#1}
\csname url@samestyle\endcsname
\providecommand{\newblock}{\relax}
\providecommand{\bibinfo}[2]{#2}
\providecommand{\BIBentrySTDinterwordspacing}{\spaceskip=0pt\relax}
\providecommand{\BIBentryALTinterwordstretchfactor}{4}
\providecommand{\BIBentryALTinterwordspacing}{\spaceskip=\fontdimen2\font plus
\BIBentryALTinterwordstretchfactor\fontdimen3\font minus
  \fontdimen4\font\relax}
\providecommand{\BIBforeignlanguage}[2]{{%
\expandafter\ifx\csname l@#1\endcsname\relax
\typeout{** WARNING: IEEEtran.bst: No hyphenation pattern has been}%
\typeout{** loaded for the language `#1'. Using the pattern for}%
\typeout{** the default language instead.}%
\else
\language=\csname l@#1\endcsname
\fi
#2}}
\providecommand{\BIBdecl}{\relax}
\BIBdecl

\bibitem{recent}
S.~Aghajanzadeh and D.~Forsyth, ``Towards robust low light image enhancement,''
  2022.

\bibitem{HA}
T.~Cohen, M.~Geiger, and M.~Weiler, ``A general theory of equivariant cnn's on
  homogeneous spaces,'' in \emph{NeurIPS}, 2019.

\bibitem{GD}
\BIBentryALTinterwordspacing
J.~E. Gerken, J.~Aronsson, O.~Carlsson, H.~Linander, F.~Ohlsson, C.~Petersson,
  and D.~Persson, ``Geometric deep learning and equivariant neural networks,''
  in \emph{arxiv}, 2021. [Online]. Available:
  \url{https://arxiv.org/abs/2105.13926}
\BIBentrySTDinterwordspacing

\bibitem{CohenWelling}
T.~S. Cohen and M.~Welling, ``Group equivariant convolutional networks,'' in
  \emph{ICML}, 2016.

\bibitem{LencVedaldi}
K.~Lenc and A.~Vedaldi, ``Understanding image representations by measuring
  their equivariance and equivalence,'' \emph{Int J Comput Vis}, vol. 127, no.
  456–476, 2019.

\bibitem{PermInv}
J.~Hartford, D.~R. Graham, K.~Leyton-Brown, and S.~Ravanbakhsh, ``Deep models
  of interactions across sets,'' in \emph{ICML}, 2018.

\bibitem{EGN}
\BIBentryALTinterwordspacing
V.~G. Satorras, E.~Hoogeboom, and M.~Welling, ``E(n) equivariant graph neural
  networks,'' in \emph{Arxiv}, 2021. [Online]. Available:
  \url{https://arxiv.org/abs/2102.09844}
\BIBentrySTDinterwordspacing

\bibitem{EPot}
S.~Batzner, A.~Musaelian, L.~Sun, M.~Geiger, J.~Mailoa, M.~Kornbluth,
  N.~Molinari, T.~Smidt, and B.~Kozinsky, ``E(3)-equivariant graph neural
  networks for data-efficient and accurate interatomic potentials,'' \emph{Nat
  Commun}, 2022.

\bibitem{ET}
N.Keriven and G.~Peyre, ``Universal invariant and equivariant graph neural
  networks,'' in \emph{NeurIPS}, 2019.

\bibitem{DAFRockTPAMI}
\BIBentryALTinterwordspacing
D.~Forsyth and J.~Rock, ``Intrinsic image decomposition using paradigms,''
  \emph{TPAMI}, In Press. [Online]. Available:
  \url{https://www.computer.org/csdl/journal/tp/5555/01/09573351/1xH5D2WNbEc}
\BIBentrySTDinterwordspacing

\bibitem{unet_2015}
O.~Ronneberger, P.~Fischer, and T.~Brox, ``U-net: Convolutional networks for
  biomedical image segmentation,'' 2015.

\bibitem{resnet_2015}
K.~He, X.~Zhang, S.~Ren, and J.~Sun, ``Deep residual learning for image
  recognition,'' 2015.

\bibitem{im2im_2018}
P.~Isola, J.-Y. Zhu, T.~Zhou, and A.~A. Efros, ``Image-to-image translation
  with conditional adversarial networks,'' 2018.

\bibitem{LOL_2018}
C.~Wei, W.~Wang, W.~Yang, and J.~Liu, ``Deep retinex decomposition for
  low-light enhancement,'' 2018.

\bibitem{deeplabv3}
L.~Chen, G.~Papandreou, F.~Schroff, and H.~Adam, ``Rethinking atrous
  convolution for semantic image segmentation,'' 2017.

\bibitem{fivek}
V.~Bychkovsky, S.~Paris, E.~Chan, and F.~Durand, ``Learning photographic global
  tonal adjustment with a database of input / output image pairs,'' in
  \emph{The Twenty-Fourth IEEE Conference on Computer Vision and Pattern
  Recognition}, 2011.

\bibitem{MBLLEN_2018}
F.~Lv, F.~Lu, J.~Wu, and C.~S. Lim, ``Mbllen: Low-light image/video enhancement
  using cnns,'' in \emph{BMVC}, 2018.

\bibitem{DSLR_2021}
S.~Lim and W.~Kim, ``Dslr: Deep stacked laplacian restorer for low-light image
  enhancement,'' \emph{IEEE Transactions on Multimedia}, vol.~23, pp.
  4272--4284, 2021.

\bibitem{LLNet_2017}
\BIBentryALTinterwordspacing
K.~G. Lore, A.~Akintayo, and S.~Sarkar, ``Llnet: {A} deep autoencoder approach
  to natural low-light image enhancement,'' \emph{PR}, vol. abs/1511.03995,
  2017. [Online]. Available: \url{http://arxiv.org/abs/1511.03995}
\BIBentrySTDinterwordspacing

\bibitem{LightenNet_2018}
C.~Li, J.~Guo, F.~Porikli, and Y.~Pang, ``Lightennet: A convolutional neural
  network for weakly illuminated image enhancement,'' \emph{PRL}, vol. 104, pp.
  15--22, 2018.

\bibitem{Retinex-Net_2018}
\BIBentryALTinterwordspacing
C.~Wei, W.~Wang, W.~Yang, and J.~Liu, ``Deep retinex decomposition for
  low-light enhancement,'' \emph{BMVC}, vol. abs/1808.04560, 2018. [Online].
  Available: \url{http://arxiv.org/abs/1808.04560}
\BIBentrySTDinterwordspacing

\bibitem{KinD_2019}
\BIBentryALTinterwordspacing
Y.~Zhang, J.~Zhang, and X.~Guo, ``Kindling the darkness: {A} practical
  low-light image enhancer,'' \emph{ACMMM}, vol. abs/1905.04161, 2019.
  [Online]. Available: \url{http://arxiv.org/abs/1905.04161}
\BIBentrySTDinterwordspacing

\bibitem{KinD++_2021}
Y.~Zhang, X.~Guo, J.~Ma, W.~Liu, and J.~Zhang, ``Beyond brightening low-light
  images,'' \emph{International Journal of Computer Vision}, pp. 1--25, 2021.

\bibitem{TBEFN_2021}
K.~Lu and L.~Zhang, ``Tbefn: A two-branch exposure-fusion network for low-light
  image enhancement,'' \emph{IEEE Transactions on Multimedia}, vol.~23, pp.
  4093--4105, 2021.

\bibitem{EnlightenGAN_2021}
Y.~fan Jiang, X.~Gong, D.~Liu, Y.~Cheng, C.~Fang, X.~Shen, J.~Yang, P.~Zhou,
  and Z.~Wang, ``Enlightengan: Deep light enhancement without paired
  supervision,'' \emph{IEEE Transactions on Image Processing}, vol.~30, pp.
  2340--2349, 2021.

\bibitem{DRBN_2020}
W.~Yang, S.~Wang, Y.~Fang, Y.~Wang, and J.~Liu, ``From fidelity to perceptual
  quality: A semi-supervised approach for low-light image enhancement,'' in
  \emph{2020 IEEE/CVF Conference on Computer Vision and Pattern Recognition
  (CVPR)}, 2020, pp. 3060--3069.

\bibitem{ExCNet_2019}
A.~Zhu, L.~Zhang, Y.~Shen, Y.~Ma, S.~Zhao, and Y.~Zhou, ``Zero-shot restoration
  of underexposed images via robust retinex decomposition,'' \emph{ACMMM}, pp.
  1--6, 2019.

\bibitem{Zero-DCE_2019}
L.~Zhang, L.~Zhang, X.~Liu, Y.~Shen, S.~Zhang, and S.~Zhao, ``Zero-shot
  restoration of back-lit images using deep internal learning,''
  \emph{Proceedings of the 27th ACM International Conference on Multimedia},
  2019.

\bibitem{RRDNet_2020}
A.~Zhu, L.~Zhang, Y.~Shen, Y.~Ma, S.~Zhao, and Y.~Zhou, ``Zero-shot restoration
  of underexposed images via robust retinex decomposition,'' \emph{2020 IEEE
  International Conference on Multimedia and Expo (ICME)}, pp. 1--6, 2020.

\bibitem{sdsd}
R.~Wang, X.~Xu, C.-W. Fu, J.~Lu, B.~Yu, and J.~Jia, ``Seeing dynamic scene in
  the dark: A high-quality video dataset with mechatronic alignment,'' in
  \emph{Proceedings of the IEEE/CVF International Conference on Computer Vision
  (ICCV)}, October 2021, pp. 9700--9709.

\bibitem{smid}
C.~Chen, Q.~Chen, M.~N. Do, and V.~Koltun, ``Seeing motion in the dark,'' in
  \emph{Proceedings of the IEEE/CVF International Conference on Computer Vision
  (ICCV)}, 2019.

\bibitem{DeepUPE_2019}
R.~Wang, Q.~Zhang, C.-W. Fu, X.~Shen, W.-S. Zheng, and J.~Jia, ``Underexposed
  photo enhancement using deep illumination estimation,'' in \emph{The IEEE
  Conference on Computer Vision and Pattern Recognition (CVPR)}, June 2019.

\bibitem{DeepLPF_2020}
S.~Moran, P.~Marza, S.~McDonagh, S.~Parisot, and G.~Slabaugh, ``Deeplpf: Deep
  local parametric filters for image enhancement,'' in \emph{Proceedings of the
  IEEE/CVF Conference on Computer Vision and Pattern Recognition (CVPR)}, June
  2020.

\bibitem{smoid}
H.~Jiang and Y.~Zheng, ``Learning-to-see-moving-objects-in-the-dark,'' in
  \emph{Proceedings of the IEEE/CVF International Conference on Computer Vision
  (ICCV)}, 2019.

\end{thebibliography}

\end{document}